\documentclass{llncs}
\usepackage{epsf}
\usepackage{graphicx}
\usepackage{cite}
\usepackage{amsmath,amssymb,amsfonts}
\usepackage{framed}
\usepackage{tabu}
\usepackage{wasysym}
\usepackage{algorithmic}
\usepackage{graphicx}
\usepackage{textcomp}
\usepackage{xcolor}
\usepackage{colortbl}
\usepackage{hyperref}
\usepackage{etoolbox}
\makeatletter
\renewcommand\section{\@startsection{section}{1}{\z@}%
                       {-8\p@ \@plus -4\p@ \@minus -4\p@}%
                       {6\p@ \@plus 4\p@ \@minus 4\p@}%
                       {\normalfont\large\bfseries\boldmath
                        \rightskip=\z@ \@plus 8em\pretolerance=10000 }}
\renewcommand\subsection{\@startsection{subsection}{2}{\z@}%
                       {-8\p@ \@plus -4\p@ \@minus -4\p@}%
                       {6\p@ \@plus 4\p@ \@minus 4\p@}%
                       {\normalfont\normalsize\bfseries\boldmath
                        \rightskip=\z@ \@plus 8em\pretolerance=10000 }}
\renewcommand\subsubsection{\@startsection{subsubsection}{3}{\z@}%
                       {-4\p@ \@plus -4\p@ \@minus -4\p@}%
                       {-1.5em \@plus -0.22em \@minus -0.1em}%
                       {\normalfont\normalsize\bfseries\boldmath}}
\makeatother

\definecolor{verylightblue}{RGB}{240,248,255}
\definecolor{lightblue}{RGB}{225,235,255}
\definecolor{blue}{RGB}{13,123,255}
\definecolor{verylightred}{RGB}{255,240,242}
\definecolor{lightred}{RGB}{255,230,230}
\definecolor{red}{RGB}{204,0,0}
\definecolor{verylightgreen}{RGB}{240,255,245}
\definecolor{lightgreen}{RGB}{230,255,238}
\definecolor{green}{RGB}{0,179,57}
\definecolor{verylightpurple}{RGB}{249,242,255}
\definecolor{lightpurple}{RGB}{229,204,255}
\definecolor{purple}{RGB}{99,0,204}
\newcommand\tab[1][1cm]{\hspace*{#1}}

\begin{document}

\author{
Alan Aipe, Mukuntha N S, Asif Ekbal}
\institute{Department of Computer Science and Engineering\\
          Indian Institute of Technology Patna\\
          Patna, India \\
          \{alan.me14,mukuntha.cs16,asif\}@iitp.ac.in}

\title{Sentiment-Aware Recommendation System for Healthcare using Social Media}

\maketitle

\begin{abstract}
Over the last decade, health communities (known as forums) have evolved into platforms where more and more users share their medical experiences, thereby seeking guidance and interacting with people of the community. The shared content, though informal and unstructured in nature, contains valuable medical and/or health related information and can be leveraged to produce structured suggestions to the common people. In this paper, at first we propose a stacked deep learning model for sentiment analysis from the medical forum data. The stacked model comprises of Convolutional Neural Network (CNN) followed by a Long Short Term Memory (LSTM) and then by another CNN. For a blog classified with positive sentiment, we retrieve the top-n similar posts. Thereafter, we develop a probabilistic model for suggesting the suitable treatments or procedures for a particular disease or health condition. We believe that integration of medical sentiment and suggestion would be beneficial to the users for finding the relevant contents regarding medications and medical conditions, without having to manually stroll through a large amount of unstructured contents.
\keywords{Health social media \and Deep learning \and Suggestion mining \and Medical sentiment}
\end{abstract}

\section{Introduction}
With the increasing popularity of electronic-bulletin boards, there has been a phenomenal growth in the amount of social media information available online. Users post about their experiences on social media such as medical forums and message-boards, seeking guidance and emotional support from the online community. As discussed in \cite{Denecke:2015:HWS:2830607}, medical social media is an increasingly viable source of useful information. These users, who are often patients themselves or the friends and/or relatives of patients write their personal views and/or experiences. Their posts are rich in information such as their experiences with disease and their satisfaction with treatment methods and diagnosis.

As discussed in \cite{DENECKE201517}, medical sentiment refers to a patient's health status, medical conditions and treatment. Extraction of this information as well as its analysis have several potential applications. The difficulty in the extraction of information such as sentiment and suggestions from a forum post can be attributed to a variety of reasons. Forum posts contain informal language combined with the usages of medical conditions and terms. The medical domain itself is sensitive to misinformation. Thus, any system built on this data would also have to incorporate relevant domain knowledge.

\subsection{Problem Definition}
\label{problemstatement}
Our main objective 
is to develop a sentiment-aware recommendation system to help build a patient assisted health-care system. We propose a novel framework for mining medical sentiments and suggestions from medical forum data. This broad objective can be modularized into the following set of research questions:
\begin{framed}\noindent
\textbf{RQ1}: Can an efficient multi-class classifier be developed to help us understand the overall medical sentiment expressed in a medical forum post?
\end{framed}
\begin{framed}\noindent
\textbf{RQ2}: How can we model the similarity between the two medical forum posts?
\end{framed}
\begin{framed}\noindent
\textbf{RQ3}: Can we propose an effective algorithm for treatment suggestion by leveraging medical sentiment obtained from the forum posts?
\end{framed}

By addressing these research questions, we aim to create a patient assisted health-care system, which is able to determine the sentiments of any user that s/he expresses in the forum post, point the user to similar forum posts for more information, and suggest possible treatment(s) or procedural methods for the user's symptoms and possible disorders.

\subsection{Motivation}
The amount of health related information being sought after on the Internet is on the rise. As discussed in \cite{pmid14728167}, an estimated 6.75 million health-related searches are made on Google every day. The Pew Internet Survey \cite{healthonline2013} claims that 35\% of U.S. adults have used the internet to diagnose a medical condition they themselves or another might have, and that 41\% of these online diagnosers have had their suspicions confirmed by a clinician. There has also been an increase in the number of health-related forums and discussion boards on the Internet, which contain useful information that is yet to be properly harnessed. Sentiment analysis has various applications. We believe it can also provide important information in health-care. In addition to doctor's advice, connecting with other people who have been in similar situations can help with several practical difficulties. According to The Pew Internet Survey, 24\% of all adults have obtained information or support from others who are having the same health conditions. A person posting on such a forum is often looking for emotional support from similar people. Consider the following two blog posts:
\begin{framed}\noindent
\textbf{Post 1}: Hi. I have been on sertaking 50mgs for about 2 months now and previously was at 25mg for two weeks. Overall my mood is alot more stable and I dont worry as much as I did before however I thought I would have a bath and when I dried my hair etc I started to feel anxious, lightheaded and all the lovely feeling you get with panic. Jus feel so yuck at the moment but my day was actually fine. This one just came out of the blue.. I wanted to know if anyone else still gets some bad moments on these. I don't know if they feel more intense as I have been feeling good for a while now. Would love to hear others stories.
\end{framed}
\begin{framed}\noindent
\textbf{Post 2}: Just wanna let you all know who are suffering with head aches/pressure that I finally went to the doctor. Told him how mines been lasting close to 6 weeks and he did a routine check up and says he's pretty I have chronic tension headaches. He prescribed me muscle relaxers, 6 visits to neck massages at a physical therapist and told me some neck exercises to do. I went in on Tuesday and since yesterday morning things have gotten better. I'm so happy I'm finally getting my life back. Just wanted you all to know so maybe you can feel better soon  
\end{framed}
In the first post, the author discusses an experience with a drug, and is looking to hear from people with similar issues. In the second post, the author discusses a positive experience and seeks to help people with similar problems.
One of our aims is to develop a system to automatically retrieve and provide such a user with the posts most similar to theirs.
Also, in order to make an informed decision knowing patient's satisfaction for a given course of treatment might be useful. We also seek to provide suggestions for treatment for a particular patient's problems. The suggestions can subsequently be verified by a qualified professional, and then be prescribed to the patients, or in more innocuous cases (such as with 'more sleep' or 'meditation'), can be directly taken as advice.

\subsection{Contributions}
In this paper, we propose a sentiment-aware patient assisted health-care system using the information extracted from the medical forums. We propose a deep learning model with a stacked architecture that makes use of Convolutional Neural Network (CNN) layers, and a Long-Short Term Memory (LSTM) network, for the classification of a blog post into its medical sentiment class. To the best of our knowledge, exploring the usage of medical sentiment to retrieve similar posts (from medical blogs) and treatment options is not yet attempted. We summarize the contributions of our proposed work as follows:
\begin{itemize}
\item We propose an effective deep learning based stacked model utilizing CNN and LSTM for medical sentiment classification.
\item We develop a method for retrieving the relevant medical forum posts, similar to a given post.
\item We propose an effective algorithm for treatment suggestions, that could lead towards building a patient care system.
\end{itemize}
\section{Related Works }

Social media is a source of huge information that can be leveraged for building many socially intelligent systems. Sentiment analysis has been explored quite extensively in various domains. However, this has not been addressed in the medical/health domain in the required measure. 

In \cite{DENECKE201517}, authors have analyzed the peculiarities of sentiment and word usage in medical forums, and performed quantitative analysis on clinical narratives and medical social media resources. In \cite{Denecke:2015:HWS:2830607}, multiple notions of sentiment analysis with reference to medical text are discussed in details. In \cite{pmid22195073}, authors have built a system that identifies drugs that cause serious adverse reactions, using messages discussing them from online health forums. They use an ensemble of Na\"{i}ve Bayes (NB) and Support Vector Machines (SVMs) classifiers to successfully identify the past drugs withdrawn from the market. Similarly, in \cite{Yang:2012:SMM:2389707.2389714}, users' written contents from social media were used to mine the association between drugs for predicting the Adverse Drug Reactions (ADRs). FDA alerts were used as gold standard, and the statistic Proportional Reporting Ratios (PRR) was shown to be of high importance in solving the problem. In \cite{zuccon2016ir}, one of the shared tasks involved the retrieval of medical forum posts related to the search queries provided. The queries involved were short, detailed and to the point, typically being less than 10 words. Our work however, focuses more on medical sentiment involved in an entire forum post, and helps to retrieve the similar posts. Recently, \cite{medicalsentiment2018} presented a benchmark setup for analyzing the medical sentiment of users on social media. They identified and analyzed multiple forms of medical sentiments in text from forum posts, and developed a corresponding annotation scheme. They also annotated and released a benchmark dataset for the problem. 

In our current work we propose a novel stacked deep learning based ensemble model for sentiment analysis in the medical domain. This is significantly different from the prior works mentioned above. 
To the best of our knowledge, no prior attempt has been made to exploit medical sentiment from social media posts to suggest treatment options to build a patient-assisted recommendation system.  %
\section{Proposed Framework}
In this section, we describe our proposed framework comprising of three phases, each of which tackles a research question enumerated in Section \ref{problemstatement}.
\subsection{Sentiment Classification}
\label{sentimentclassification}
\begin{figure*}[ht]
   \begin{center}
   \includegraphics[width=1.00\columnwidth]{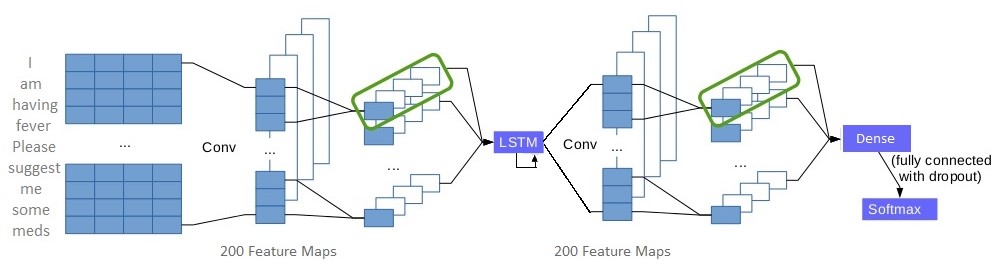}
   \caption{Stacked CNN-LSTM-CNN Ensemble Architecture for medical sentiment classification}
   \label{fig:cnnlstmcnn}
   \end{center}
\end{figure*}

Medical sentiment refers to analyzing the health status reflected in a given social media post. We approach this task as a multi-class classification problem using sentiment taxonomy as described in Section \ref{sentimenttaxonomy}. Convolutional Neural Network (CNN) architectures have been extensively applied to sentiment analysis and classification tasks \cite{CNN:2014, medicalsentiment2018}. Long Short Term Memory (LSTMs) are a special kind of Recurrent Neural Network (RNN) capable of learning long-term dependencies by handling the vanishing and exploding gradient problem \cite{LSTM:1997}. We propose an architecture consisting of two deep Convolutional Neural Network (CNN) layers and a Long-Short Term Memory (LSTM) layer, stacked with a fully connected layer followed by three-neurons output layer having softmax as an activation function. A diagrammatic representation of the classifier is shown in Fig. \ref{fig:cnnlstmcnn}. The social media posts are first vectorized (as discussed in Section \ref{textvectorization}) and then fed as input to the classifier. Convolutional layers, used in the classifier, generate 200 dimensional feature maps of unigram and bigram filter sizes. Feature maps from the final CNN layer are maxpooled, flattened and fed into the fully connected layer having a rectified linear unit (ReLU) activation function. The output of the above mentioned layer is fed into another fully connected layer with a softmax activation to obtain class probabilities. Sentiment denoted by the class having the highest softmax value is considered to be the medical sentiment of the input message. The intuition behind adopting a CNN-LSTM-CNN architecture is as follows: During close scrutiny of the dataset, we observed that users often share experiences adhering to their time frame. For example, ``I was suffering from anxiety. My doctor asked me to take cit 20mg per day. Now I feel better". In this post, the user portrays his/her initial condition, explains the treatment which was used and then the effect of the treatment- all in a very timely sequence. Moreover, health status also keeps changing in the same sequence. This trend was observed throughout the dataset. Therefore, temporal features are the key to medical sentiment classification. Hence, in our stacked ensemble model, first CNN layer extracts top-level features, then LSTM finds the temporal relationships between the extracted features and the final CNN layer filters out the top temporal relationships which are subsequently fed into a fully connected layer.
\subsection{Top-N Similar Posts Retrieval}
\label{topnsimilarpost}
Users often share contents in forums, seeking guidance and to connect with other people who experienced similar medical scenarios. Thus, retrieving top-N similar posts would help to focus on contents which are relevant to one's medical condition, without having to manually scan through all the forum posts. We could have posed this task as a regression problem where a machine learning (ML) model learns to predict similarity score for a given pair of forum posts, but there is no suitable dataset available for this task to the best of our knowledge.
We tackle this task by creating a similarity metric (as shown in E.q. \ref{overallsim}) and evaluating it by manually annotating a small test set (as discussed in Section \ref{topnevaluation}). The similarity metric comprises of three terms:
\begin{itemize}
\item \textit{Disease Similarity}: It refers to the Jaccard similarity score computed between the two forum posts with respect to the diseases mentioned in the posts. Section \ref{umlsconcept} discusses how the diseases are extracted from a given post. Let \textit{J(A,B)} denotes the Jaccard similarity between set A and B, \textit{DS(P,Q)} denotes the disease similarity between two forum posts P and Q, \textit{D(P)} and \textit{D(Q)} denote the set of diseases mentioned in P and Q respectively, then:
\begin{equation}
DS(P,Q) = J(D(P),D(Q))
\end{equation}
where,
\begin{eqnarray}\nonumber
J(A,B) = \frac{|A \cap B|}{|A \cup B|}
\end{eqnarray}
\item \textit{Symptom similarity} : It refers to the Jaccard similarity between two forum posts with respect to the symptoms mentioned in them. Section \ref{umlsconcept} discusses how the symptoms mentioned in texts are extracted from a given post. Let \textit{SS(P,Q)} denotes the disease similarity between two forum posts P and Q, \textit{S(P)} and \textit{S(Q)} denote the set of diseases mentioned in P and Q respectively, then
\begin{equation}
SS(P,Q) = J(S(P),S(Q))
\end{equation}
\item \textit{Text similarity} : It refers to the cosine similarity between the document vectors corresponding to two forum posts. Document vector of a post is the sum of vectors of all the words (Section \ref{textvectorization}) in a given sentence. Let $\vec{D_P}$ and $\vec{D_Q}$ denote the document vectors corresponding to the forum posts P and Q, \textit{TS(P,Q)} denotes the cosine similarity between them, then
\begin{equation}
TS(P,Q) = \dfrac{\vec{D_P} \cdot \vec{D_Q}}{|\vec{D_P}|\times|\vec{D_Q}|}
\end{equation}
\end{itemize}
 We compute the above similarities between a pair of posts, and use Equation \ref{overallsim} to obtain the overall similarity score \textit{Sim(P,Q)} between two given forum posts P and Q. For a given test instance, training posts are ranked according to the similarity score (with respect to test) and top-N posts are retrieved.
 \begin{eqnarray}
 MISim(P,Q) = \dfrac{2 \times DS(P,Q) + SS(P,Q)}{3}\\
 \label{overallsim}
 Sim(P,Q) = \dfrac{2 \times MISim(P,Q) + TS(P,Q)}{3}
\end{eqnarray}
\tab where \textit{MISim(P,Q)} denotes the similarity between P and Q with respect to the relevant medical information.\newline\newline
The main objective of similar post retrieval is to search for the posts depicting similar medical experience. Medical information shared in a forum post can be considered as an aggregate of the disease conditions and symptoms encountered. Medical experience shared in a forum post can be considered as an aggregate of the medical information shared and the semantic meaning of the text, in the same order of relevance.
This is the intuition behind adoption of the similarity metric in Equation \ref{overallsim}.
\subsection{Treatment Suggestion}
\label{treatmentsuggestion}
A treatment T mentioned in a forum post P can be considered suitable for a disease D mentioned in post Q if P and Q depict similar medical experience and the probability that T to produce a positive medical sentiment, given D. Thus, suggestion score \textit{G(T,D)} is given by,
\begin{equation}
\label{treatmentsuggestionformula}
G(T,D) = Sim(P,Q) \times Pr(+veSentiment | T,D)
\end{equation}
\begin{eqnarray}\nonumber
G(T,D)&\geq&\tau \text{\tab Treatment T is suggested}\\ \nonumber
  &<& \tau \text{\tab Treatment T is not suggested}
\end{eqnarray}
\tab where $\tau$ is a hyper-parameter of the framework and \textit{Pr(A)} denotes the probability of event A. %
\section{Dataset and Experimental Setup}
In this section, we discuss the details of the datasets used for our experiments and the evaluation setups.
\subsection{Forum dataset}
\label{dataset}
We perform experiments using a recently released datasets for sentiment analysis\cite{medicalsentiment2018}. This dataset consists of social media posts collected from the medical forum 'patient.info'. In total 5,189 posts were segregated into three classes -- Exist, Recover, Deteriorate based on medical conditions the post described, and 3,675 posts were classified into three classes -- Effective, Ineffective, Serious Adverse Effect based on the effect of medication. As our framework operates at a generic level, we combine both the segments into a single dataset, mapping labels from each segment to a sentiment taxonomy as discussed in Section \ref{sentimenttaxonomy}. The classes with respect to medical condition are redefined as follows:
\begin{itemize}
\item \textbf{Exist}: User shares the symptoms of any medical problem. This is mapped to the \textit{neutral sentiment}.
\item \textbf{Recover}: Users share their recovery status from the previous problems. This is mapped to the \textit{positive sentiment}.
\item \textbf{Deteriorate}: User share information about their worsening health conditions. We map this to the \textit{negative sentiment}.
\end{itemize}
The classes with respect to the effect of medication are:
\begin{itemize}
\item \textbf{Effective}: User shares information about the usefulness of treatment. This is mapped to the \textit{positive sentiment}.
\item \textbf{Ineffective}: User shares information that the treatment undergone has no effect as such. These are mapped to the \textit{neutral sentiment}.
\item \textbf{Serious adverse effect}: User shares negative opinions towards the treatment, mainly due to adverse drug effect. This is mapped to the \textit{negative sentiment}.
\end{itemize}
\subsubsection{Sentiment Taxonomy}
\label{sentimenttaxonomy}
A different sentiment taxonomy is conceptualized keeping in mind the generic behavior of our proposed system. It does not distinguish between the forum posts related to medical conditions and medication. Thus, a one-to-one mapping from sentiment classes used in each segment of the dataset to a more generic taxonomy is essential. We show the class distribution in Table \ref{table:classdist}.
\begin{table}[t!]
\begin{center}
\begin{tabular}{m{3.5cm} lccc}
\rowcolor{red}
\hline 
\rowcolor{blue}
\textcolor{white}{\bf Sentiment} &   \textcolor{white}{\bf Distribution(\%)}\\ \hline
\rowcolor{lightblue}
Positive & 37.49\\
\rowcolor{verylightblue}
Neutral & 32.34\\
\rowcolor{lightblue}
Negative & 30.17 \\
\hline
\end{tabular}
\end{center}
\caption{Class distribution in the dataset with respect to sentiment taxonomy}
\label{table:classdist}
\end{table}
\begin{itemize}
\item \textbf{Positive sentiment} : Forum posts depicting improvement in overall health status or positive results of the treatment. \newline
For example : "I have been suffering from anxiety for a couple of years. Yesterday, my doc prescribed me Xanax. I am feeling better now." This post is considered positive as it depicts positive results of Xanax.
\item \textbf{Negative sentiment} : Forum posts describing deteriorating  health status or negative results of treatment.\newline
For example : "Can citalopram make it really hard for you to sleep? i cant sleep i feel wide awake every night for the last week and im on it for 7 weeks."%
\item \textbf{Neutral sentiment}: This denotes to the forum posts where neither positive nor negative sentiment is expressed, with no change in overall health status of the person.
\newline
For example : "I was wondering if anyone has used Xanax for anxiety and stress. I have a doctors appointment tomorrow and not sure what will be decided to use."
\end{itemize}
\subsection{Word Embeddings}
\label{wordembeddings}
Capturing semantic similarity between the target texts is an important step towards accurate classification. For this reason, word embeddings play a pivotal role. We use the pre-trained word2vec \cite{word2vec} model\footnote{http://bio.nlplab.org/}, induced from the PubMed and PMC texts along with the texts extracted from the Wikipedia dump.
\subsection{Tools used and Preprocessing}
The codebase, during experimentation, is written in Python (version 3.6) with external libraries -- namely \emph{keras}\footnote{https://keras.io/} for neural network design, \emph{sklearn}\footnote{http://scikit-learn.org/} for evaluation of baseline and the proposed model, \emph{pandas}\footnote{http://pandas.pydata.org/} for easier access of data in the form of tables (or, data frames) during execution, \emph{nltk}\footnote{http://www.nltk.org/} for textual analysis and \emph{pickle}\footnote{https://docs.python.org/3/library/pickle.html} for saving and retrieving input-output of different modules from the secondary storage devices. The preprocessing phase comprises of the removal of non-ASCII characters, stop words and handling of non alphanumeric characters followed by tokenization. Tokens of size (number of characters) less than 3 were also removed due to a very low probability of these becoming indicative features to the classification model. Labels corresponding to sentiment classes of each segment in the dataset are mapped to the generic taxonomy classes (as discussed in Section \ref{sentimenttaxonomy}), and the corresponding one-hot encodings are generated.
\subsubsection{Text vectorization}
\label{textvectorization}
Using the pre-trained word2vec model (discussed in Section \ref{wordembeddings}), each token is converted to a 200-dimensional vector. They are stacked together and padded to form a 2-D matrix of desired size (150 x 200). The number 150 denotes the maximum number of tokens in any preprocessed forum posts belonging to the training set. 
\subsection{UMLS Concept Retrieval}
\label{umlsconcept}
Identification of medical information like diseases, symptoms and treatments mentioned in a forum post is essential for the top-n similar post retrieval (Section \ref{topnsimilarpost}) and treatment suggestion (Section \ref{treatmentsuggestion}) phases of the proposed framework. The Unified Medical Language System\footnote{https://www.nlm.nih.gov/research/umls/} (UMLS) is a compendium of many controlled vocabularies in the biomedical sciences (created in 1986). Therefore UMLS concept identifiers, related to the above mentioned medical information, were retrieved using Apache cTAKES\footnote{ctakes.apache.org/}. Concepts with semantic type 'Disorder or Disease' were added to the set of diseases-those with semantic types 'Sign or Symptom' were added to the set of symptoms and those with semantic types 'Medication' and 'Procedures' were added to the set of treatments. 
\subsection{Relevance judgement for similar post retrieval}
\label{topnevaluation}
Annotating pairs of forum posts with their similarity scores as per human judgment is necessary to evaluate how much the retrieved text is relevant. This corresponds to the evaluation of the proposed custom similarity metric (E.q. \ref{overallsim}). Since annotating every pairs of posts is a cumbersome task, 20\% of the total posts in the dataset were randomly selected, maintaining equal class distribution for the annotation purpose. For each such post, top 5 similar posts are retrieved using the similarity metric. Annotators were asked to judge the similarity between each retrieved post and the original post on a Likert-type scale, from 1 to 5 (1 represents high dissimilarity while 5 represents high similarity between a pair of posts). Annotators were provided with the guidelines for relevance judgments on two questions--`Is this post relevant to the original post in terms of medical information?' and `Are the experiences and situations depicted in the pair of posts similar?'. A pair of posts is given high similarity rating if both of the conditions are true, and a low rating if neither is true. Three annotators having post-graduate educational levels performed the annotations.

We measure the inter-annotator agreement using Krippendorff's alpha metric\cite{krippendorff2011computing}, and this was observed to be 0.78. Disagreements between the annotators can be explained on the basis of ambiguities, encountered during the labeling task. We provide few examples below: 
\begin{enumerate}
\item There are cases where original writer of the blog assigned higher rating (denoting relevant), but the annotator disagreed on what constituted a `relevant' post. This often corresponds to the posts giving general advice for illness. For example,'You can take xanax in case of high stress. It worked for me.' Such advice may not be applicable to a certain specific situation.
\item Ambiguities are also observed for the cases where the authors of the posts are of similar age, sex and socio-economic backgrounds, but have different health issues (for example, one post depicted a male teenager with severe health anxiety, while the other post described a male teenager with social anxiety). For such cases, similarity ratings were varied.
\item Ratings also vary in cases where the symptoms match, but the cause and disorder differ. Annotators face problem in judging the posts which do not contain enough medical information. For example, headache can be a symptom for different diseases.%
\end{enumerate} %
\section{Experimental Results and Analysis}
In this section, we report the evaluation results and present necessary analysis.
\subsection{Sentiment Classification}
\label{sec:classification}
\begin{table*}[b]
\begin{center}
\begin{tabular}{lllccc}
\rowcolor{green}
\hline 
\textcolor{white}{\bf Model} &\textcolor{white}{\bf Accuracy} &   \textcolor{white}{\bf Cohen-Kappa} & \multicolumn{3}{c}{\textcolor{white}{\bf Macro}} 
\\ \hline
\rowcolor{verylightgreen}
 & & & Precision & Recall & F1-Score \\
 \rowcolor{lightgreen}
 Baseline \cite{medicalsentiment2018} & 0.63 & 0.443 & 0.661 & 0.643 & 0.652 \\
 \rowcolor{verylightgreen}
LSTM & 0.609 & 0.411 & 0.632 & 0.628 & 0.63 \\
\rowcolor{lightgreen}
CNN-LSTM & 0.6516 & 0.4559 & 0.6846 & 0.6604 & 0.6708 \\
\rowcolor{verylightgreen}
Proposed model &\bf 0.6919 & \bf 0.4966 & \bf 0.7179 & \bf 0.7002 & \bf 0.7089 \\
\hline
\end{tabular}
\end{center}
\caption{Evaluation results of 5-fold cross-validation for sentiment classification.}
\label{table:results1}
\end{table*}
The classification model (described in Section \ref{sentimentclassification}) is trained on a dataset of 8,864 unique instances obtained after preprocessing. We define a baseline model by implementing the CNN based system as proposed in \cite{medicalsentiment2018} under the identical experimental conditions as that of our proposed architecture. We also develop a model based on an LSTM. To see the real impact of the third layer, we also show the performance of a CNN-LSTM based model. Batch size for training was set to 32. Results of 5-fold cross validation are shown in Table \ref{table:results1}.

Evaluation shows that the proposed model performs better than the baseline system, and efficiently captures medical sentiment from the social media posts. Table  \ref{table:results1} shows the  accuracy, cohen-kappa, precision, recall and F1 score of the proposed model as  0.6919, 0.4966, 0.7179, 0.7002 and 0.7089, respectively. In comparison to the baseline model this is approximately a 9.13\% improvement in terms of all the metrics. Posts usually consist of medical events and experiences. Therefore, capturing temporally related spatially close features is required for inferring the overall health status. The proposed CNN-LSTM-CNN network has been shown to be better at this task compared to the other models.

The high value of the Cohen-Kappa metric suggests that the proposed model indeed learns to classify posts into 3 sentiment classes rather than making any random guess.
A closer look at the classification errors revealed that there are instances where CNN and LSTM predict incorrectly, but the proposed model correctly classifies. With the following example where baseline and LSTM both failed to correctly classify, but the proposed model succeeded:
`I had a doctors appointment today. He told I was recovering and should be more optimistic. I am still anxious and stressed most of the time'. Baseline model and LSTM classified it as positive (might be because of terms like 'recovering' and 'optimistic') while the proposed model classified it as negative. This shows that the proposed model can satisfactorily capture the contextual information, and leverage it effectively for the classification task. 

\begin{figure}[t]
   \begin{center}
   \includegraphics[width=0.3\columnwidth]{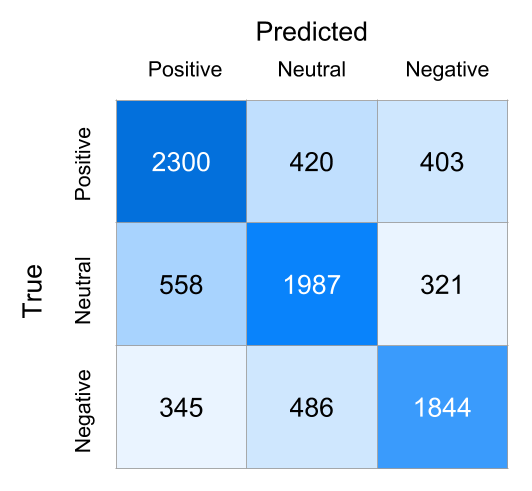}
   \caption{Confusion matrix of sentiment classification}
   \label{fig:confusionmatrix}
   \end{center}
\end{figure}

To understand where our system fails we perform detailed error analysis-both quantitatively and qualitatively. We show quantitative analysis in terms of confusion matrix as shown in Figure \ref{fig:confusionmatrix}.
Close scrutiny of the predicted and actual values of test instances reveals that majority of misclassification occurs in cases where sentiment remains positive or negative throughout the post, and suddenly alters at the end of the post. For example: ``I have been suffering from anxiety for a couple of years now. Doctors kept prescribing new and new medicines but I felt no change. Stress was unbearable. I lost my parents last year. The grief made me even worse. But I am currently feeling good after taking coQ10". 
We observe that the proposed model was confused in such cases.
Moreover, users often share some personal content which does not help the medical domain significantly. Such noises also contribute to the misclassification.

\textbf{Comparison with existing models:}
One of the recent works on medical sentiment analysis is reported in  \cite{medicalsentiment2018}.
They have trained and evaluated a CNN based architecture separately for medical condition and medication segments.
As discussed in the dataset and experiment section, we have merged both the datasets related to medical condition and medications into one for training and evaluation.
Our definition of medical sentiment is, thus, more generic in nature, and direct comparison to the existing system is not very rational. None of the related works mentioned in the related works section addressed sentiment analysis for medical suggestion mining. The experimental setups, datasets and the sentiment classes used in all these works are also very different. 

\subsection{Top-N similar post retrieval}
Evaluation of the retrieval task is done by comparing similarity scores assigned for a pair of forum posts by the system and by human annotator (as discussed in Section \ref{topnevaluation}). Our focus is to determine the correlation between the similarity score assigned to the pairs of posts through human and the system judgments (rather than on actual similarity values). That is if a human feels that a post P is more relavant to post Q than post R, then the system also operates in the same way. Therefore, we use pearson correlation coefficient for the evaluation purpose. Statistical significance of the correlation (2-tailed p-value from \textit{T-test} with null hypothesis that the correlation occured by chance) was found to be 0.00139, 0.0344 and 0.0186, respectively, for each sentiment class. Precision@5 is also calculated to evaluate the relevance of the retrieved forum posts. As annotation was done using top-5 retrieved posts (as discussed in Section \ref{topnevaluation}), Recall@5 could not be calculated. 

We design a baseline model using K-nearest neighbour algorithm that makes use of cosine similarity metric for capturing the textual similarity. We show the results in Table \ref{table:topnpc}. From the evaluation results, it is evident that similarity scores assigned by the proposed system are more positively correlated with the human judgments than the baseline. Correlation can be considered statistically significant as the p-values corresponding to all the sentiment classes are less than 0.05. The better Precision@5 metric corresponds that a greater number of relevant posts are retrieved by the proposed approach in comparison to the baseline model. 

We also calculate the Discounted Cumulative Gain (DCG) \cite{Jarvelin:2002:CGE:582415.582418} of the similarity scores for both models from the human judgments. The idea behind DCG is that highly relevant documents appearing lower in the ranking should be penalized. A logarithmic reduction factor was applied to the human relevance judgment which was scaled from 0 to 4, and the DCG accumulated at a rank position 5 was calculated with the following formula: 
\begin{equation}
DCG_{5} = \sum_{i=1}^{5} \frac{rel_{i}}{\log_{2}(i+1)}
\end{equation}
where $rel_{i}$ is the relevance judgment of the post at position $i$. The NDCG could not be calculated, as annotation was done only using top-5 retrieved posts (as discussed in Section \ref{topnevaluation}).

During error analysis, we observe few forum posts where users share their personal feelings, but due to the presence of less medically relevant contents, these are labeled as irrelevant by the system. However, these contain some relevant information that could be useful to the end users. 
For example, 'Hello everyone, Good morning to all. I know I had been away for a couple of days. I went outing with my family to get away from the stress I had been feeling lately. Strolled thorugh the park, played tennis with kids and visited cool places nearby. U know Family is the best therapy for every problem. Still feeling a little bit anxious lately. Suggest me something' -- The example blog contains proportionally more personal information than the medically relevant one. 
However, 'Feeling a little bit anxious lately' is the medically relevant part of the post. Thus, filtering out such contents is required for better performance and would help the system to focus better on the relevant contents. 
There are two possible ways to tackle this problem. We would like to look into these in future. 
\begin{enumerate}
\item Increasing the weight of medical information similarity (represented as \textit{MISim} in Equation 5) while computing the overall similarity score. 
\item Identifying and removing personal, medically irrelevant contents using (possibly) by either designing a sequence labeling model (classifying relevant vs. irrelevant) or by manually verifying the data or by finding certain phrases or snippets from the blog. 
\end{enumerate} 
\begin{table*}[hbt]
\begin{center}
\begin{tabular}{lcccccc}
\rowcolor{red}
\hline 
\textcolor{white}{\bf Sentiment} &
\multicolumn{2}{c}{\textcolor{white}{\bf Pearson correlation}} &
\multicolumn{2}{c}{\textcolor{white}{\bf Precision@5}} & 
\multicolumn{2}{c}{\textcolor{white}{\bf DCG$_{\text{5}}$}}
\\ 
\hline
\rowcolor{lightred}
 & A & B & A & B & A & B\\
\rowcolor{verylightred}
Positive &\bf 0.3586 & 0.2104 &\bf 0.6638 & 0.5467 & \bf 6.0106 & 2.1826\\
\rowcolor{lightred}
Neutral &  \bf 0.3297 & 0.2748 & \bf 0.5932 & 0.5734 & \bf 5.3541 & 2.4353\\
\rowcolor{verylightred}
Negative &\bf 0.3345 & 0.2334 & \bf 0.623 & 0.5321 & \bf 4.7361 & 2.4825\\
\hline
\end{tabular}
\end{center}
\caption{Evaluation corresponding to the top-n similar posts retrieval. 'A' and 'B' denote the results corresponding to the proposed metric and K-nearest neighbor algorithm using text similarity metric, respectively. }
\label{table:topnpc}
\end{table*}
\subsection{Treatment suggestion}
Evaluation of treatment suggestion is particularly challenging because it requires the annotators with high level of medical expertise. Moreover to the best of our knowledge there is no existing benchmark dataset for this evaluation.
Hence, we are not able to provide any quantitative evaluation of the suggestion module. However, it is to be noted that our suggestion module is based on the soundness of sentiment classification module. Our evaluation presented in the earlier section shows that our sentiment classifier has acceptable output quality. 
The task of a good treatment suggestion system is to mine the best and relevant treatment suggestion for a candidate disease. As the function for computing the suggestion score (Eq. \ref{treatmentsuggestionformula}) involves computing the probability of positive sentiment, given a treatment T and disorder/disease D, it is always ensured that T is a candidate treatment for D, i.e. the treatment T produced positive results in context of D in at least one case. In other words, the probability term ensures that irrelevant treatments that did not give positive result in context of D would never appear as treatment suggestion for D.
The efficiency of the suggestion module depends on the following three factors: 
\begin{enumerate}
\item%
Apache cTAKES retrieved correct concepts in majority of the cases with only a few exceptions, which are mostly ambiguous in nature. For example, the word 'basis' can represent clinically-proven NAD+ Supplement or can be used as synonym of the word 'premise'.

\item%
If an irrelevant post is labeled as relevant by the system, then suggestions shouldn't contain treatments mentioned in that post.
Thus, the similarity metric plays an important role in picking the right treatment for a given candidate disease.

\item Value of hyper-parameter $\tau$ (E.q. \ref{treatmentsuggestionformula}): As its value decreases, more number of candidate treatments are suggested by the system.
\end{enumerate}
Performance of the module can be augmented and tailored by tweaking the above parameters depending on the practical application in hand.
\section{Conclusion and Future Work}
In this paper, we have established the usefulness of medical sentiment analysis for building a recommendation system that will assist building a patient assisted health-care system. A deep learning model has been presented for classifying the medical sentiment expressed in a forum post into conventional polarity-based classes. We have empirically shown that the proposed architecture can satisfactorily capture sentiment from the social media posts. We have also proposed a novel similarity metric for the retrieval of forum posts with similar medical experiences and sentiments. A novel treatment suggestion algorithm has been also proposed, that utilizes our similarity metric along with the patient-treatment satisfaction ratings. We have performed a very detailed analysis of our model.

 In our work, we use the UMLS database due to its wide usage and acceptability as a standard database.%
 We also point to other future work, such as annotating a dataset for treatment suggestions -- which would increase the scope of machine learning, developing a sequence labeling model to remove personal irrelevant contents etc. Our work serves as an initial study in harnessing the huge amounts of open, useful information available on medical forums.

\section{Acknowledgements}
Asif Ekbal acknowledges the Young Faculty Research Fellowship (YFRF), supported by Visvesvaraya PhD scheme for Electronics and IT, Ministry of Electronics and Information Technology (MeitY), Government of India, being implemented by Digital India Corporation (formerly Media Lab Asia).  
\bibliographystyle{splncs04}
\bibliography{main}
\end{document}